\documentclass[11pt,a4paper]{article}
\usepackage[hyperref]{naaclhlt2018}
\usepackage{arydshln}
\usepackage{graphicx}
\usepackage{latexsym}
\usepackage{listings}
\usepackage{multirow}
\usepackage{times}
\usepackage{url}

\usepackage{xspace}
\usepackage{subcaption}
\usepackage{booktabs}
\usepackage{amssymb}
\DeclareCaptionFont{10pt}{\fontsize{10pt}{12pt}\selectfont}
\aclfinalcopy % Uncomment this line for the final submission
\def\aclpaperid{***} %  Enter the acl Paper ID

\newcommand{\mysubsection}[1]{\vspace{0.3em} \noindent\textbf{#1}}

\title{Factors Influencing the Surprising Instability of Word Embeddings}
\author{Laura Wendlandt, Jonathan K. Kummerfeld \and Rada Mihalcea \\
  Computer Science \& Engineering \\
  University of Michigan, Ann Arbor \\
  {\tt \{wenlaura,jkummerf,mihalcea\}@umich.edu}}

\date{}

\begin{document}
\maketitle
\begin{abstract}
    Despite the recent popularity of word embedding methods, there is only a small body of work exploring the limitations of these representations. In this paper, we consider one aspect of embedding spaces, namely their stability.
    We show that even relatively high frequency words (100-200 occurrences) are often unstable.
    We provide empirical evidence for how various factors contribute to the stability of word embeddings, and we analyze the effects of stability on downstream tasks.
\end{abstract}

\section{Introduction}

Word embeddings are low-dimensional, dense vector representations that capture semantic properties of words. Recently, they have gained tremendous popularity in Natural Language Processing (NLP) and have been used in tasks as diverse as text similarity \cite{kenter2015short}, part-of-speech tagging \cite{tsvetkov2016learning}, sentiment analysis \cite{faruqui2014retrofitting}, and machine translation \cite{mikolov2013exploiting}. Although word embeddings are widely used across NLP, their stability has not yet been fully evaluated and understood. In this paper, we explore the factors that play a role in the stability of word embeddings, including properties of the data, properties of the algorithm, and properties of the words. We find that word embeddings exhibit substantial instabilities, which can have implications for downstream tasks. 

\begin{figure}[!t]
    \centering
    \includegraphics[width=0.5\textwidth]{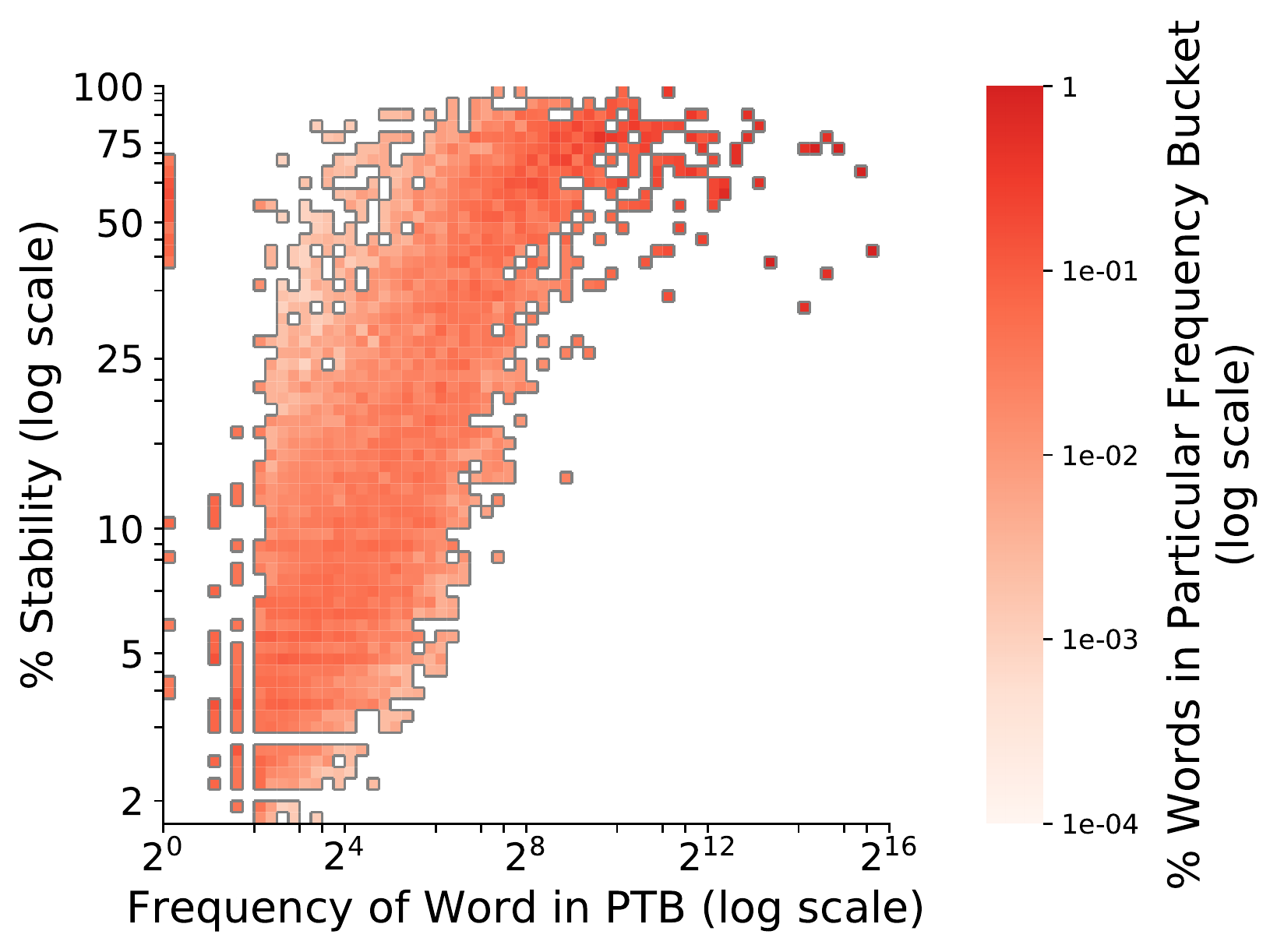}
    \caption{Stability of \textit{word2vec} as a property of frequency in the PTB. Stability is measured across ten randomized embedding spaces trained on the training portion of the PTB (determined using language modeling splits \cite{mikolov2010recurrent}). Each word is placed in a frequency bucket (x-axis), and each column (frequency bucket) is normalized.}
    \label{fig:ptb_stability}
\end{figure}

Using the overlap between nearest neighbors in an embedding space as a measure of stability (see \autoref{sec:definingStability} below for more information), we observe that many common embedding spaces have large amounts of instability. For example, Figure~\ref{fig:ptb_stability} shows the instability of the embeddings obtained by training \textit{word2vec} on the Penn Treebank (PTB) \cite{marcus1993building}. As expected, lower frequency words have lower stability and higher frequency words have higher stability. What is surprising however about this graph is the medium-frequency words, which show huge variance in stability. This cannot be explained by frequency, so there must be other factors contributing to their instability.

In the following experiments, we explore which factors affect stability, as well as how this stability affects downstream tasks that word embeddings are commonly used for. To our knowledge, this is the first study comprehensively examining the factors behind instability.

\section{Related Work}

There has been much recent interest in the applications of word embeddings, as well as a small, but growing, amount of work analyzing the properties of word embeddings.

Here, we explore three different embedding methods: PPMI \cite{bullinaria2007extracting}, \textit{word2vec} \cite{mikolov2013distributed}, and GloVe \cite{pennington2014glove}. Various aspects of the embedding spaces produced by these algorithms have been previously studied. Particularly, the effect of parameter choices has a large impact on how all three of these algorithms behave 
\cite{levy2015improving}. Further work shows that the parameters of the embedding algorithm \textit{word2vec} influence the geometry of word vectors and their context vectors \cite{mimno2017strange}. These parameters can be optimized; Hellrich and Hahn (\citeyear{hellrich2016bad}) posit optimal parameters for negative sampling and the number of epochs to train for. They also demonstrate that in addition to parameter settings, word properties, such as word ambiguity, affect embedding quality.

In addition to exploring word and algorithmic parameters, concurrent work by Antoniak and Mimno (\citeyear{antoniakevaluating}) evaluates how document properties affect the stability of word embeddings. 
We also explore the stability of embeddings, but focus on a broader range of factors, and consider the effect of stability on downstream tasks.
In contrast, Antoniak and Mimno focus on using word embeddings to analyze language \cite[e.g.,][]{Garg201720347}, rather than to perform tasks.

At a higher level of granularity, Tan et al. (\citeyear{tan2015lexical}) analyze word embedding spaces by comparing two spaces. They do this by linearly transforming one space into another space, and they show that words have different usage properties in different domains (in their case, Twitter and Wikipedia).

Finally, embeddings can be analyzed using second-order properties of embeddings (e.g., how a word relates to the words around it). Newman-Griffis and Fosler-Lussier (\citeyear{newman2017second}) validate the usefulness of second-order properties, by demonstrating that embeddings based on second-order properties perform as well as the typical first-order embeddings. Here, we use second-order properties of embeddings to quantify stability.

\section{Defining Stability}
\label{sec:definingStability}

We define {\it stability} as the percent overlap between nearest neighbors in an embedding space.\footnote{This metric is concurrently explored in work by Antoniak and Mimno (\citeyear{antoniakevaluating}).}
Given a word $W$ and two embedding spaces $A$ and $B$, take the ten nearest neighbors of $W$ in both $A$ and $B$. Let the stability of $W$ be the percent overlap between these two lists of nearest neighbors. 100\% stability indicates perfect agreement between the two embedding spaces, while 0\% stability indicates complete disagreement. In order to find the ten nearest neighbors of a word $W$ in an embedding space $A$, we measure distance between words using cosine similarity.\footnote{We found comparable results for other distance metrics, such as $l^1$ norm, $l^2$ norm, and $l^\infty$ norm, but we report results from cosine similarity to be consistent with other word embedding research.} This definition of stability can be generalized to more than two embedding spaces by considering the average overlap between two sets of embedding spaces. Let $X$ and $Y$ be two sets of embedding spaces. Then, for every pair of embedding spaces $(x,y)$, where $x\in X$ and $y\in Y$, take the ten nearest neighbors of $W$ in both $x$ and $y$ and calculate percent overlap. Let the stability be the average percent overlap over every pair of embedding spaces $(x,y)$.

Consider an example using this metric. Table~\ref{tab:international} shows the top ten nearest neighbors for the word \textit{international} in three randomly initialized \textit{word2vec} embedding spaces trained on the NYT Arts domain (see Section \ref{sec:data} for a description of this corpus). These models share some similar words, such as \textit{metropolitan} and \textit{national}, but there are also many differences. On average, each pair of models has four out of ten words in common, so the stability of \textit{international} across these three models is 40\%.

\begin{table}[tb]
    \centering
    \begin{tabular}{l | l | l}
    \hline
    Model 1 & Model 2 & Model 3 \\
    \hline
    \textbf{metropolitan} & \textit{ballet} & \textbf{national} \\
    \textbf{national} & \textbf{metropolitan} & \textit{ballet} \\
    \textit{egyptian} & bard & \textbf{metropolitan} \\
    \textit{rhode} & chicago & institute \\
    \textit{society} & \textbf{national} & glimmerglass \\
    debut & \textit{state} & \textit{egyptian} \\
    folk & \textit{exhibitions} & intensive \\
    reinstallation & \textit{society} & jazz \\
    chairwoman & whitney & \textit{state} \\
    philadelphia & \textit{rhode} & \textit{exhibitions} \\
    \hline
    \end{tabular}
    \caption{Top ten most similar words for the word \textit{international} in three randomly intialized \textit{word2vec} models trained on the NYT Arts Domain. Words in all three lists are in bold; words in only two of the lists are italicized.}\label{tab:international}
\end{table}

\begin{figure}[!htb]
    \centering
    \includegraphics[trim={0 5mm 0 0},width=0.5\textwidth]{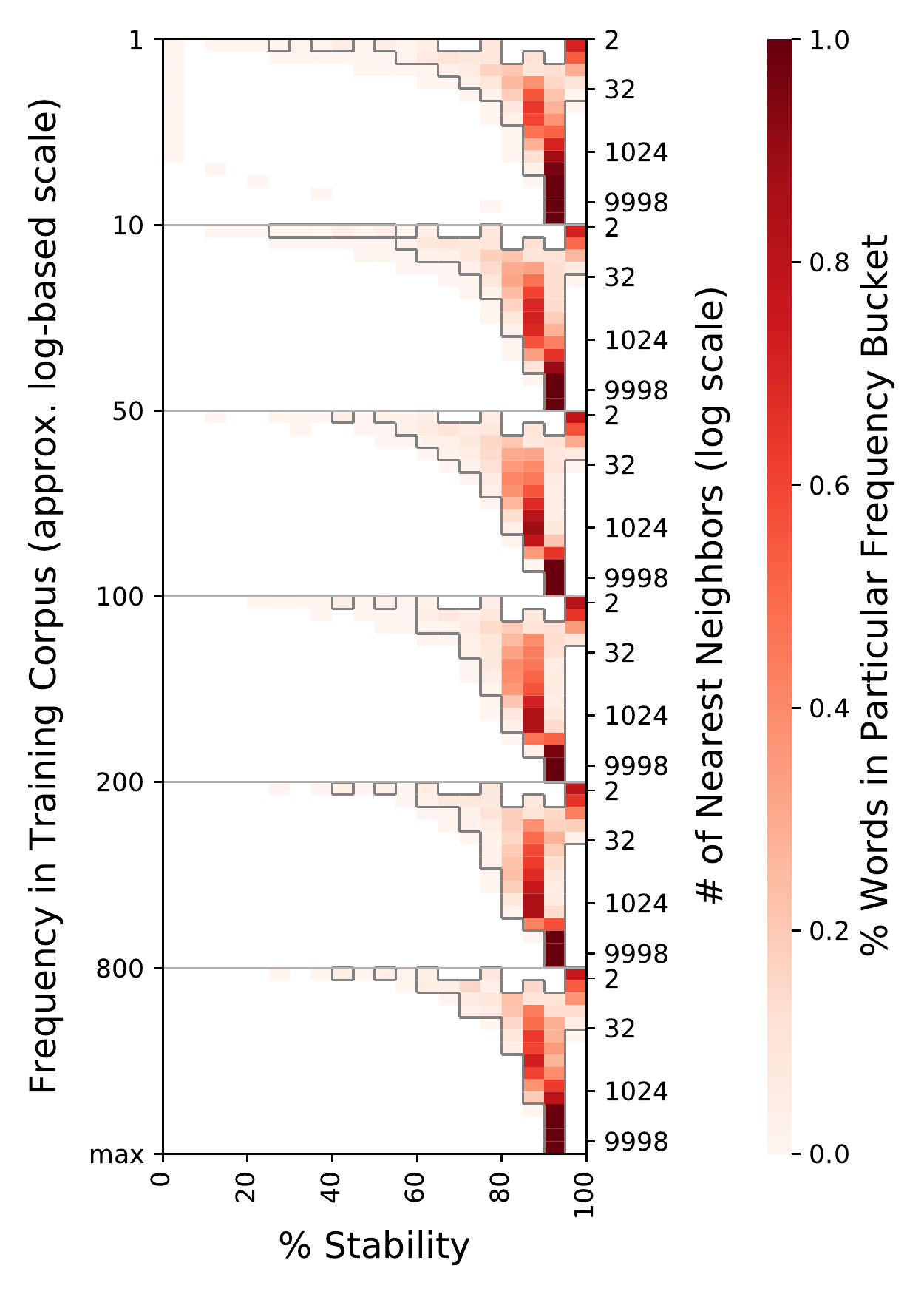}
    \caption{Stability of GloVe on the PTB. Stability is measured across ten randomized embedding spaces trained on the training data of the PTB (determined using language modeling splits \cite{mikolov2010recurrent}). Each word is placed in a frequency bucket (left y-axis) and stability is determined using a varying number of nearest neighbors for each frequency bucket (right y-axis). Each row is normalized, and boxes with more than 0.01 of the row's mass are outlined.}
    \label{fig:increasingNeighbors_glove}
\end{figure}

The idea of evaluating ten best options is also found in other tasks, like lexical substitution \cite[e.g.,][]{mccarthy2007semeval} and word association \cite[e.g.,][]{Garimella17Demograhic}, where the top ten results are considered in the final evaluation metric. To give some intuition for how changing the number of nearest neighbors affects our stability metric, consider Figure~\ref{fig:increasingNeighbors_glove}. This graph shows how the stability of GloVe changes with the frequency of the word and the number of neighbors used to calculate stability; please see the figure caption for a more detailed explanation of how this graph is structured.
Within each frequency bucket, the stability is consistent across varying numbers of neighbors. Ten nearest neighbors performs approximately as well as a higher number of nearest neighbors (e.g., 100). We see this pattern for low frequency words as well as for high frequency words. Because the performance does not change substantially by increasing the number of nearest neighbors, it is computationally less intensive to use a small number of nearest neighbors. We choose ten nearest neighbors as our metric throughout the rest of the paper.

\section{Factors Influencing Stability}

As we saw in Figure \ref{fig:ptb_stability}, embeddings are sometimes surprisingly unstable.
To understand the factors behind the (in)stability of word embeddings, we build a regression model that aims to predict the stability of a word given: (1) properties related to the word itself; (2) properties of the data used to train the embeddings; and (3) properties of the algorithm used to construct these embeddings. Using this regression model, we draw observations about factors that play a role in the stability of word embeddings.

\subsection{Methodology}

We use ridge regression to model these various factors \cite{hoerl1970ridge}. Ridge regression regularizes the magnitude of the model weights, producing a more interpretable model than non-regularized linear regression. This regularization mitigates the effects of multicollinearity (when two features are highly correlated). Specifically, given $N$ ground-truth data points with $M$ extracted features per data point, let $\mathbf{x}_n\in\mathbb{R}^{1\times M}$ be the features for sample $n$ and let $\mathbf{y}\in\mathbb{R}^{1\times N}$ be the set of labels. Then, ridge regression learns a set of weights $\mathbf{w}\in\mathbb{R}^{1\times M}$ by minimizing the least squares function with $l^2$ regularization, where $\lambda$ is a regularization constant:
$$L(\mathbf{w}) = \frac{1}{2}\sum_{n=1}^N (\mathbf{y}_n-\mathbf{w}^\top \mathbf{x}_n)^2 + \frac{\lambda}{2}||\mathbf{w}||^2$$
We set $\lambda=1$. In addition to ridge regression, we tried non-regularized linear regression. We obtained comparable results, but many of the weights were very large or very small, making them hard to interpret.

The goodness of fit of a regression model is measured using the coefficient of determination $R^2$. This measures how much variance in the dependent variable $\mathbf{y}$ is captured by the independent variables $\mathbf{x}$. A model that always predicts the expected value of $\mathbf{y}$, regardless of the input features, will receive an $R^2$ score of 0. The highest possible $R^2$ score is 1, and the $R^2$ score can be negative.

Given this model, we create training instances by observing the stability of a large number of words across various combinations of two embedding spaces. Specifically, given a word $W$ and two embedding spaces $A$ and $B$, we encode properties of the {\it word $W$}, as well as properties of the {\it datasets} and the {\it algorithms} used to train the embedding spaces $A$ and $B$. The target value associated with this features is the stability of the word $W$ across embedding spaces $A$ and $B$. We repeat this process for more than 2,500 words, several datasets, and three embedding algorithms. 

Specifically, we consider all the words present in all seven of the data domains we are using (see Section \ref{sec:data}), 2,521 words in total. Using the feature categories described below, we generate a feature vector for each unique word, dataset, algorithm, and dimension size, resulting in a total of 27,794,025 training instances. To get good average estimates for each embedding algorithm, we train each embedding space five times, randomized differently each time (this does not apply to PPMI, which has no random component). We then train a ridge regression model on these instances.
The model is trained to predict the stability of word $W$ across embedding spaces $A$ and $B$ (where $A$ and $B$ are not necessarily trained using the same algorithm, parameters, or training data). Because we are using this model to learn associations between certain features and stability, no test data is necessary. The emphasis is on the model itself, not on the model's performance on a specific task.

\begin{table}[tb]
    \centering
    \scalebox{0.9}{
        \begin{tabular}{l | l}
        \hline
        \multicolumn{2}{l}{\textbf{Word Properties}} \\
        \hline
        Primary part-of-speech & \texttt{Adjective} \\
        Secondary part-of-speech & \texttt{Noun} \\
        \# Parts-of-speech & \texttt{2} \\
        \# WordNet senses & \texttt{3} \\
        \# Syllables & \texttt{5} \\
        \hline
        \multicolumn{2}{l}{\textbf{Data Properties}} \\
        \hline
        Raw frequency in corpus $A$ & \texttt{106} \\
        Raw frequency in corpus $B$ & \texttt{669} \\
        Diff. in raw frequency & \texttt{563} \\
        Vocab. size of corpus $A$ & \texttt{10,508} \\
        Vocab. size of corpus $B$ & \texttt{43,888} \\
        Diff. in vocab. size & \texttt{33,380} \\
        Overlap in corpora vocab. & \texttt{17\%} \\
        Domains present & \texttt{Arts, Europarl} \\
        Do the domains match? & \texttt{False} \\
        Training position in $A$ & \texttt{1.02\%} \\
        Training position in $B$ & \texttt{0.15\%} \\
        Diff. in training position & \texttt{0.86\%} \\
        \hline
        \multicolumn{2}{l}{\textbf{Algorithm Properties}} \\
        \hline
        Algorithms present & \texttt{\textit{word2vec}, PPMI} \\
        Do the algorithms match? & \texttt{False} \\
        Embedding dimension of $A$ & \texttt{100} \\
        Embedding dimension of $B$ & \texttt{100} \\
        Diff. in dimension & \texttt{0} \\
        Do the dimensions match? & \texttt{True} \\
        \hline
        \end{tabular}
        }
    \caption{Consider the word \textit{international} in two embedding spaces. Suppose embedding space $A$ is trained using \textit{word2vec} (embedding dimension 100) on the NYT Arts domain, and embedding space $B$ is trained using PPMI (embedding dimension 100) on Europarl. This table summarizes the resulting features for this word across the two embedding spaces.}\label{tab:exampleFeatures}
\end{table}

We describe next each of the three main categories of factors examined in the model. An example of these features is given in Table \ref{tab:exampleFeatures}.

\subsection{Word Properties} 

We encode several features that capture attributes of the word $W$. First, we use the primary and secondary part-of-speech (POS) of the word. Both of these are represented as bags-of-words of all possible POS, and are determined by looking at the primary (most frequent) and secondary (second most frequent) POS of the word in the Brown corpus\footnote{Here, we use the universal tagset, which consists of twelve possible POS: adjective, adposition, adverb, conjunction, determiner / article, noun, numeral, particle, pronoun, verb, punctuation mark, and other \cite{petrov2011universal}.} \cite{francis1979brown}. If the word is not present in the Brown corpus, then all of these POS features are set to zero. 

To get a coarse-grained representation of the polysemy of the word, we consider the number of different POS present. For a finer-grained representation, we use the number of different WordNet senses associated with the word \cite{miller1995wordnet,fellbaum1998wordnet}. 

We also consider the number of syllables in a word, determined using the CMU Pronuncing Dictionary \cite{weide1998cmu}. If the word is not present in the dictionary, then this is set to zero.

\begin{table}[tb]
    \centering
    \scalebox{0.9}{
        \begin{tabular}{l | r | r  | r }
        \hline 
                &  & Vocab. & Num. Tokens /\\
        Dataset & Sentences &  \multicolumn{1}{c|}{Size} &  \multicolumn{1}{c}{Vocab. Size} \\
        \hline
        NYT US & 13,923 & 5,787 & 64.37 \\
        NYT NY & 36,792 & 11,182 & 80.41 \\
        NYT Business & 21,048 & 7,212 & 75.96 \\
        NYT Arts & 28,161 & 10,508 & 65.29 \\
        NYT Sports & 21,610 & 5,967 & 77.85 \\
        All NYT & 121,534 & 24,144 & 117.98 \\
        Europarl & 2,297,621 & 43,888 & 1,394.28 \\
        \hline
        \end{tabular}
        }
    \caption{Dataset statistics.}\label{tab:datasets}
\end{table}

\subsection{Data Properties}\label{sec:data}

Data features capture properties of the training data (and the word in relation to the training data). For this model, we gather data from two sources: New York Times (NYT) \cite{sandhaus2008new} and Europarl \cite{koehn2005europarl}. Overall, we consider seven domains of data: (1) NYT - U.S., (2) NYT - New York and Region, (3) NYT - Business, (4) NYT - Arts, (5) NYT - Sports, (6) All of the data from domains 1-5 (denoted ``All NYT"), and (7) All of English Europarl. Table~\ref{tab:datasets} shows statistics about these datasets. The first five domains are chosen because they are the top five most common categories of news articles present in the NYT corpus. They are smaller than ``All NYT" and Europarl, and they have a narrow topical focus. The ``All NYT" domain is more diverse topically and larger than the first five domains. Finally, the Europarl domain is the largest domain, and it is focused on a single topic (European Parliamentary politics). These varying datasets allow us to consider how data-dependent properties affect stability.

We use several features related to domain.
First, we consider the raw frequency of word $W$ in both the domain of data used for embedding space $A$ and the domain of data for space $B$. To make our regression model symmetric, we effectively encode three features: the higher raw frequency (between the two), the lower raw frequency, and the absolute difference in raw frequency. 

We also consider the vocabulary size of each corpus (again, symmetrically) and the percent overlap between corpora vocabulary, as well as the domain of each of the two corpora, represented as a bag-of-words of domains. Finally, we consider whether the two corpora are from the same domain.

Our final data-level features explore the role of curriculum learning in stability. It has been posited that the order of the training data affects the performance of certain algorithms, and previous work has shown that for some neural network-based tasks, a good training data order (curriculum learning strategy) can improve performance \cite{bengio2009curriculum}. Curriculum learning has been previously explored for \textit{word2vec}, where it has been found that optimizing training data order can lead to small improvements on common NLP tasks \cite{tsvetkov2016learning}. Of the embedding algorithms we consider, curriculum learning only affects \textit{word2vec}. Because GloVe and PPMI use the data to learn a complete matrix before building embeddings, the order of the training data will not affect their performance. To measure the effects of training data order, we include as features the first appearance of word $W$ in the dataset for embedding space $A$ and the first appearance of $W$ in the dataset for embedding space $B$ (represented as percentages of the total number of training sentences)\footnote{All \textit{word2vec} experiments reported here are run in a multi-core setting, which means that these percentages are approximate. However, comparable results were achieved using a single-core version of \textit{word2vec}.}. We further include the absolute difference between these percentages.

\subsection{Algorithm Properties}

In addition to word and data properties, we encode features about the embedding algorithms. These include the different algorithms being used, as well as the different parameter settings of these algorithms.  Here, we consider three embedding algorithms, \textit{word2vec}, GloVe, and PPMI. The choice of algorithm is represented in our feature vector as a bag-of-words.

PPMI creates embeddings by first building a positive pointwise mutual information word-context matrix, and then reducing the dimensionality of this matrix using SVD \cite{bullinaria2007extracting}. A more recent word embedding algorithm, \textit{word2vec} (skip-gram model) \cite{mikolov2013distributed} uses a shallow neural network to learn word embeddings by predicting context words. Another recent method for creating word embeddings, GloVe, is based on factoring a matrix of ratios of co-occurrence probabilities \cite{pennington2014glove}.

For each algorithm, we choose common parameter settings.
For \textit{word2vec}, two of the parameters that need to be chosen are window size and minimum count. Window size refers to the maximum distance between the current word and the predicted word (e.g., how many neighboring words to consider for each target word). Any word appearing less than the minimum count number of times in the corpus is discarded and not considered in the \textit{word2vec} algorithm. For both of these features, we choose standard parameter settings, namely, a window size of 5 and a minimum count of 5.
For GloVe, we also choose standard parameters. We use 50 iterations of the algorithm for embedding dimensions less than 300, and 100 iterations for higher dimensions.

We also add a feature reflecting the embedding dimension, namely one of five embedding dimensions: 50, 100, 200, 400, or 800.

\begin{table}[tb]
    \centering
    \scalebox{0.9}{
        \begin{tabular}{l | r}
        \hline
        Feature & Weight \\
        \hline
        Lower training data position of word $W$ & -1.52 \\
        Higher training data position of $W$ & -1.49 \\
        Primary POS = Numeral & 1.12 \\
        Primary POS = Other & -1.08 \\
        Primary POS = Punctuation mark & -1.02 \\
        Overlap between corpora vocab. & 1.01 \\
        Primary POS = Adjective & -0.92 \\
        Primary POS = Adposition & -0.92 \\
        Do the two domains match? & 0.91 \\
        Primary POS = Verb & -0.88 \\
        Primary POS = Conjunction & -0.84 \\
        Primary POS = Noun & -0.81 \\
        Primary POS = Adverb & -0.79 \\
        Do the two algorithms match? & 0.78 \\
        Secondary POS = Pronoun & 0.62 \\
        Primary POS = Determiner & -0.48 \\
        Primary POS = Particle & -0.44 \\
        Secondary POS = Particle & 0.36 \\
        Secondary POS = Other & 0.28 \\
        Primary POS = Pronoun & -0.26 \\
        Secondary POS = Verb & -0.25 \\
        Number of \textit{word2vec} embeddings & -0.21 \\
        Secondary POS = Adverb & 0.15 \\
        Secondary POS = Determiner & 0.14 \\
        Number of NYT Arts Domain & -0.14 \\
        Number of NYT All Domain & 0.14 \\
        Number of GloVe embeddings & 0.13 \\
        Number of syllables & -0.11 \\
        %Secondary POS = Conjunction & -0.08 \\
        %Number of PPMI embeddings & 0.07 \\
        %Number of different WordNet POS & 0.07 \\
        %Do the two dimensions match? & 0.07 \\
        %Number of Europarl Domain & -0.05 \\
        %Secondary POS = Adjective & -0.04 \\
        %Secondary POS = Noun & -0.04 \\
        %Absolute diff. in training data position & 0.04 \\
        %Secondary POS = Adposition & -0.03 \\
        %Number of Business Domain & 0.03 \\
        %Number of NY Domain & 0.02 \\
        \hline
        \end{tabular}
        }
    \caption{Regression weights with a magnitude greater than 0.1, sorted by magnitude.}\label{tab:metaClassifierWeights}
\end{table}

%\begin{table}[tb]
%    \centering
%    \scalebox{0.9}{
%        \begin{tabular}{l | r}
%        \hline
%        Feature & Weight \\
%        \hline
%        Lower training data position of word $W$ & -1.48 \\
%        Higher training data position of $W$ & -1.47 \\
%Primary POS = Other & -1.08 \\
%Primary POS = Punctuation & -1.02 \\
%Primary POS = Adjective & -0.924 \\
%Primary POS = Adposition & -0.919 \\
%Primary POS = Verb & -0.885 \\
%Primary POS = Conjunction & -0.846 \\
%Primary POS = Noun & -0.816 \\
%Primary POS = Adverb & -0.790 \\
%Primary POS = Determiner & -0.474 \\
%Primary POS = Particle & -0.449 \\
%Primary POS = Pronoun & -0.256 \\
%Secondary POS = Verb & -0.253 \\
%Number of \textit{word2vec} embeddings & -0.183 \\
%Number of NYT Arts Domain & -0.133 \\
%Number of syllables & -0.110 \\
%Secondary POS = Conjunction & -0.079 \\
%Number of Europarl Domain & -0.056 \\
%Secondary POS = Adjective & -0.045 \\
%Secondary POS = Noun & -0.038 \\
%Secondary POS = Adposition & -0.037 \\
%        \hline
%        \end{tabular}
%        }
%    \caption{Updated model - Regression weights with a magnitude greater than 0.1, sorted by magnitude.}\label{tab:metaClassifierWeights}
%\end{table}

\begin{figure}[!htb]
    \centering
    \begin{subfigure}[b]{0.5\textwidth}
       \includegraphics[width=1\linewidth]{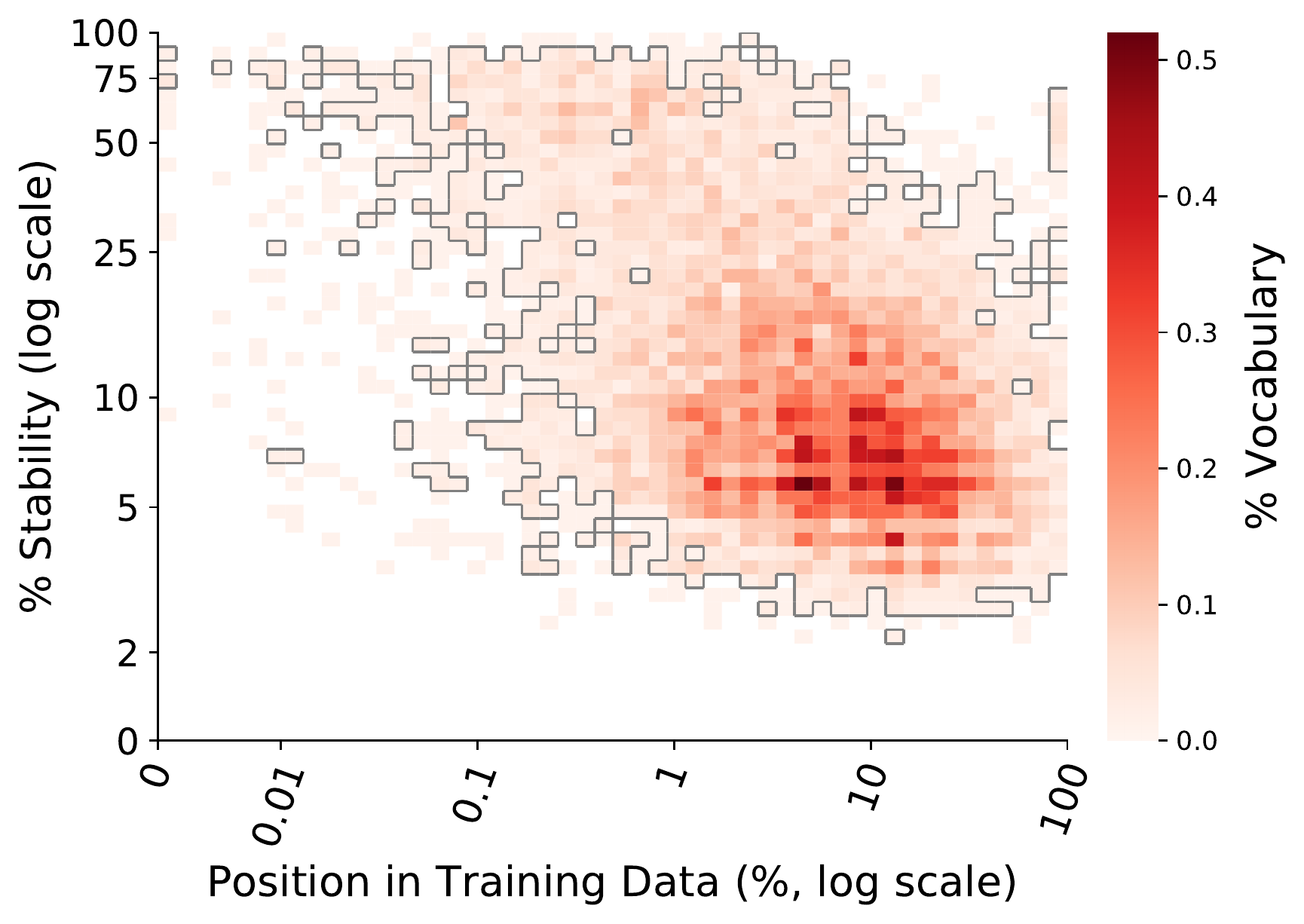}
       \caption{\textit{word2vec}}
    \end{subfigure}
    
    \begin{subfigure}[b]{0.5\textwidth}
       \includegraphics[width=1\linewidth]{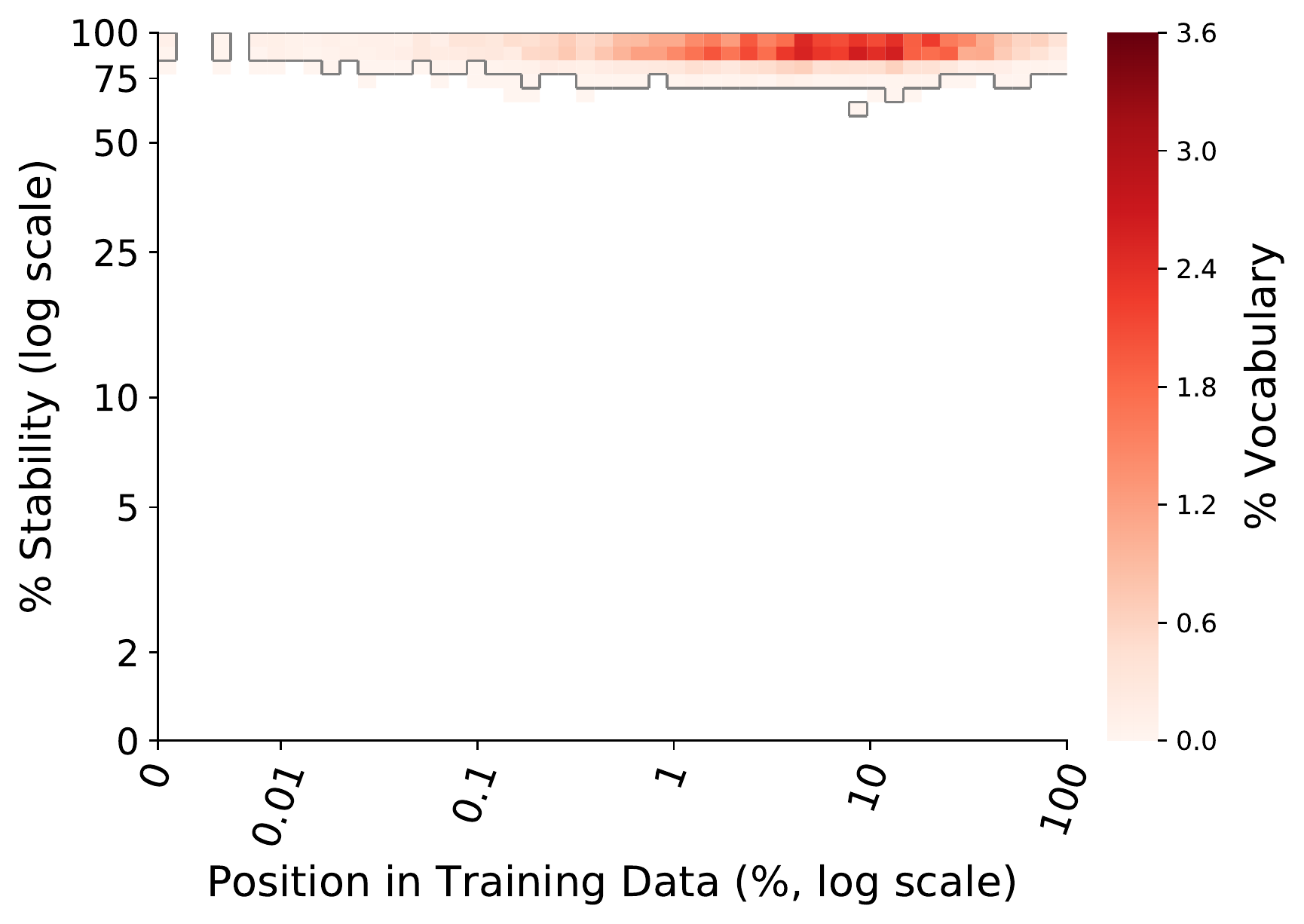}
       \caption{GloVe}
    \end{subfigure}
    
    \caption{Stability of both \textit{word2vec} and GloVe as properties of the starting word position in the training data of the PTB. Stability is measured across ten randomized embedding spaces trained on the training data of the PTB (determined using language modeling splits \cite{mikolov2010recurrent}). Boxes with more than 0.02\% of the total vocabulary mass are outlined.}
    \label{fig:ptb_stability_curriculumLearning}
\end{figure}

\section{Lessons Learned: What Contributes to the Stability of an Embedding}

Overall, the regression model achieves a coefficient of determination ($R^2$) score of 0.301 on the training data, which indicates that the regression has learned a linear model that reasonably fits the training data given.
Using the regression model, we can analyze the  weights corresponding to each of the features being considered, shown in Table~\ref{tab:metaClassifierWeights}. These weights are difficult to interpret, because features have different distributions and ranges. However, we make several general observations about the stability of word embeddings.

\mysubsection{Observation 1. Curriculum learning is important.} This is evident because the top two features (by magnitude) of the regression model capture where the word first appears in the training data. Figure~\ref{fig:ptb_stability_curriculumLearning} shows trends between training data position and stability in the PTB. This figure contrasts \textit{word2vec} with GloVe (which is order invariant).

To further understand the effect of curriculum learning on the model, we train a regression model with all of the features except the curriculum learning features. This model achieves an $R^2$ score of 0.291 (compared to the full model's score of 0.301). This indicates that curriculum learning is a factor in stability.

\begin{table}[tb]
    \centering
    \begin{tabular}{l | c}
    \hline
    Primary POS & Avg. Stability \\
    \hline
    Numeral & 47\% \\
    Verb & 31\% \\
    Determiner & 31\% \\
    Adjective & 31\% \\
    Noun & 30\% \\
    Adverb & 29\% \\
    Pronoun & 29\% \\
    Conjunction & 28\% \\
    Particle & 26\% \\
    Adposition & 25\% \\
    Punctuation mark & 22\% \\
    \hline
    \end{tabular}
    \caption{Percent stability broken down by part-of-speech, ordered by decreasing stability.}\label{tab:pos_stabilities}
\end{table}

\mysubsection{Observation 2. POS is one of the biggest factors in stability.} Table~\ref{tab:metaClassifierWeights} shows that many of the top weights belong to POS-related features (both primary and secondary POS). Table~\ref{tab:pos_stabilities} compares average stabilities for each primary POS. Here we see that the most stable POS are numerals, verbs, and determiners, while the least stable POS are punctuation marks, adpositions, and particles.

\begin{figure}[!htb]
    \centering
    \includegraphics[width=0.5\textwidth]{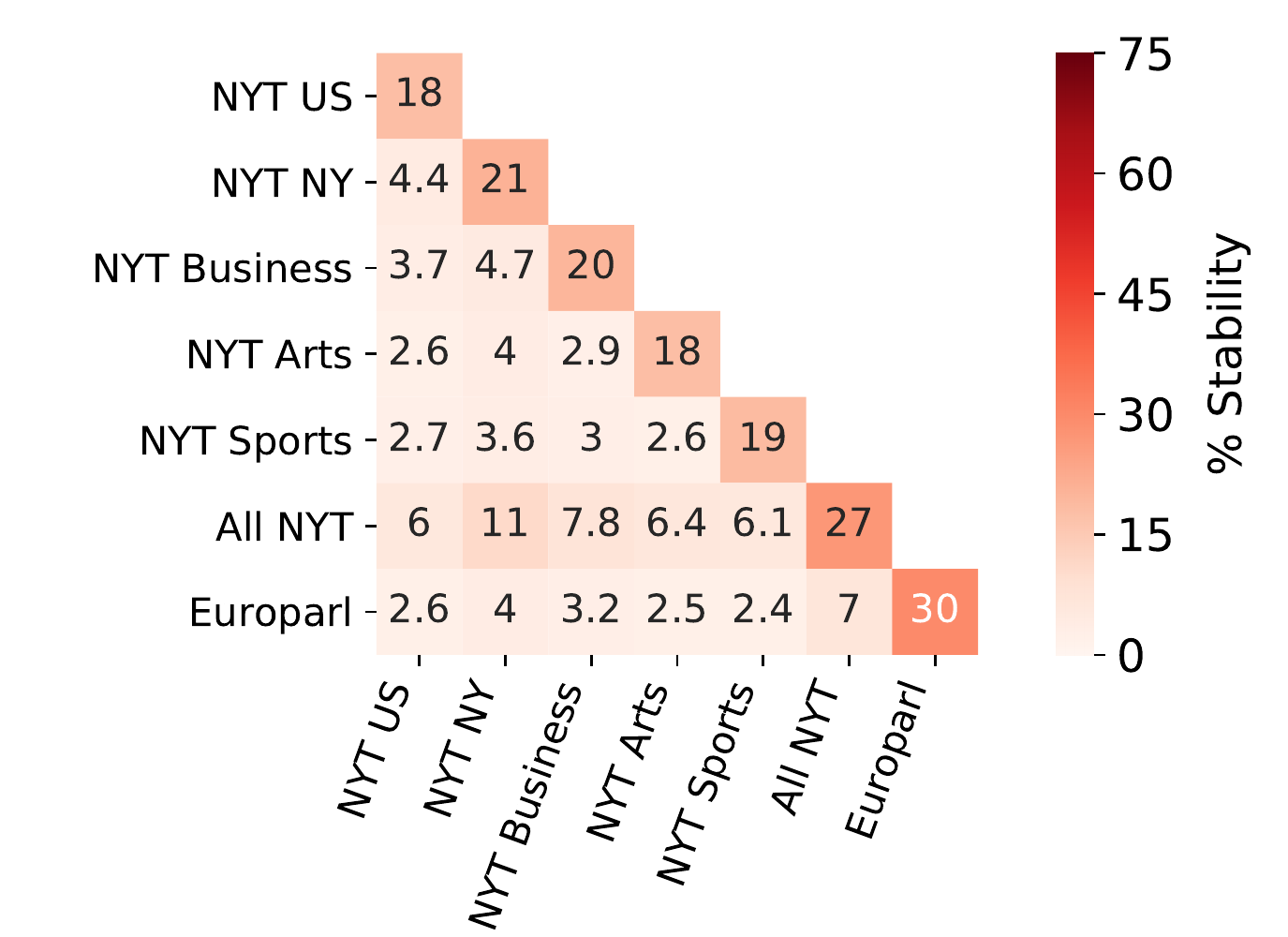}
    \caption{Percent stability broken down by domain.}
    \label{fig:domain}
    
    \vspace{0.5cm}
    \includegraphics[width=0.3\textwidth]{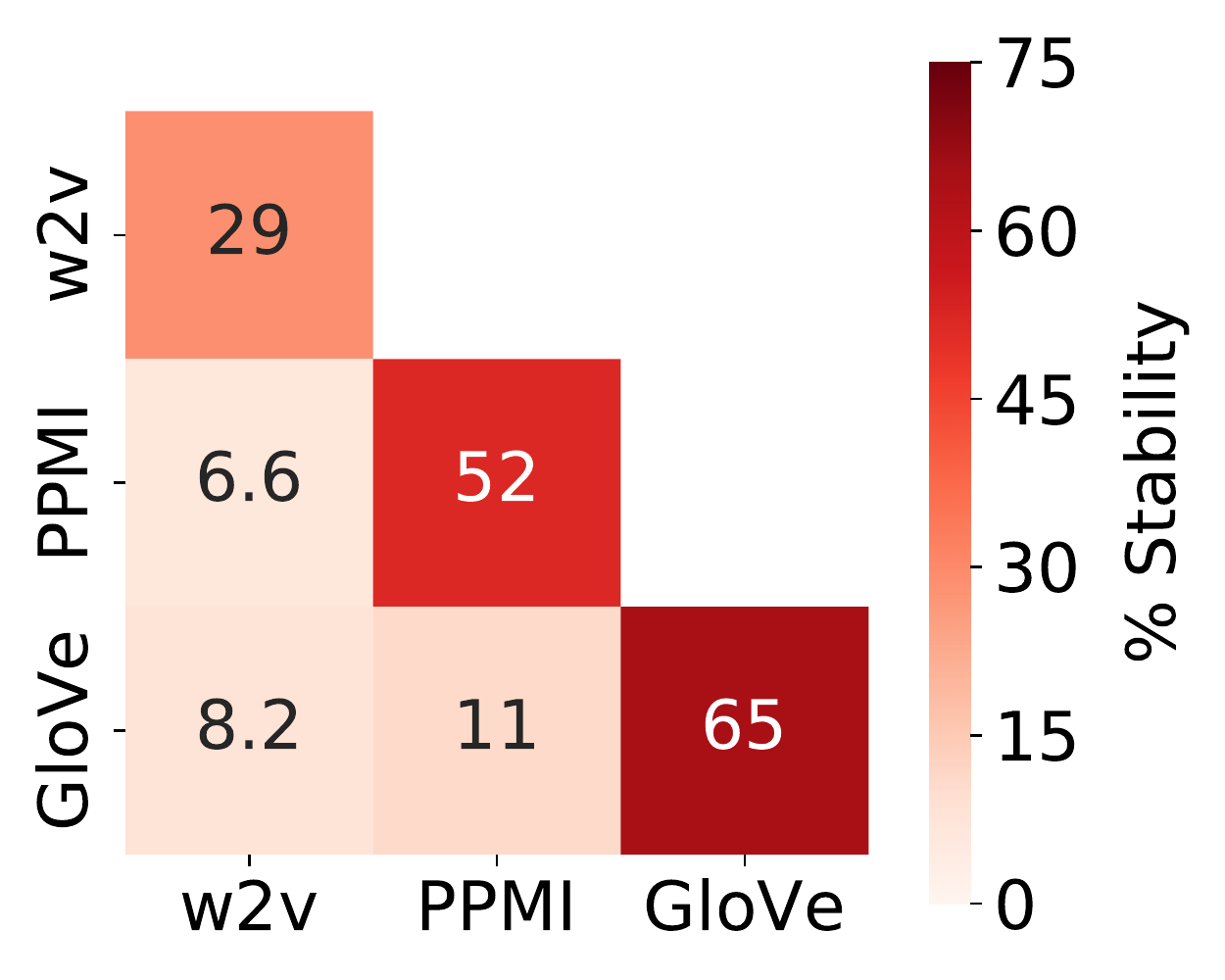}
    \caption{Percent stability broken down between algorithm (in-domain data only).}
    \label{fig:in_domain_algorithms}
\end{figure}

\mysubsection{Observation 3. Stability within domains is greater than stability across domains.} Table~\ref{tab:metaClassifierWeights} shows that many of the top factors are domain-related. Figure~\ref{fig:domain} shows the results of the regression model broken down by domain. This figure shows the highest stabilities appearing on the diagonal of the matrix, where the two embedding spaces both belong to the same domain. The stabilities are substantially lower off the diagonal.

Figure~\ref{fig:domain} also shows that ``All NYT" generalizes across the other NYT domains better than Europarl, but not as well as in-domain data (``All NYT" encompasses data from US, NY, Business, Arts, and Sports). This is true even though Europarl is much larger than ``All NYT".

\mysubsection{Observation 4. Overall, GloVe is the most stable embedding algorithm.} This is particularly apparent when only in-domain data is considered, as in Figure~\ref{fig:in_domain_algorithms}. PPMI achieves similar stability, while \textit{word2vec} lags considerably behind.

\begin{figure}[!htb]
    \centering
    \includegraphics[width=0.5\textwidth]{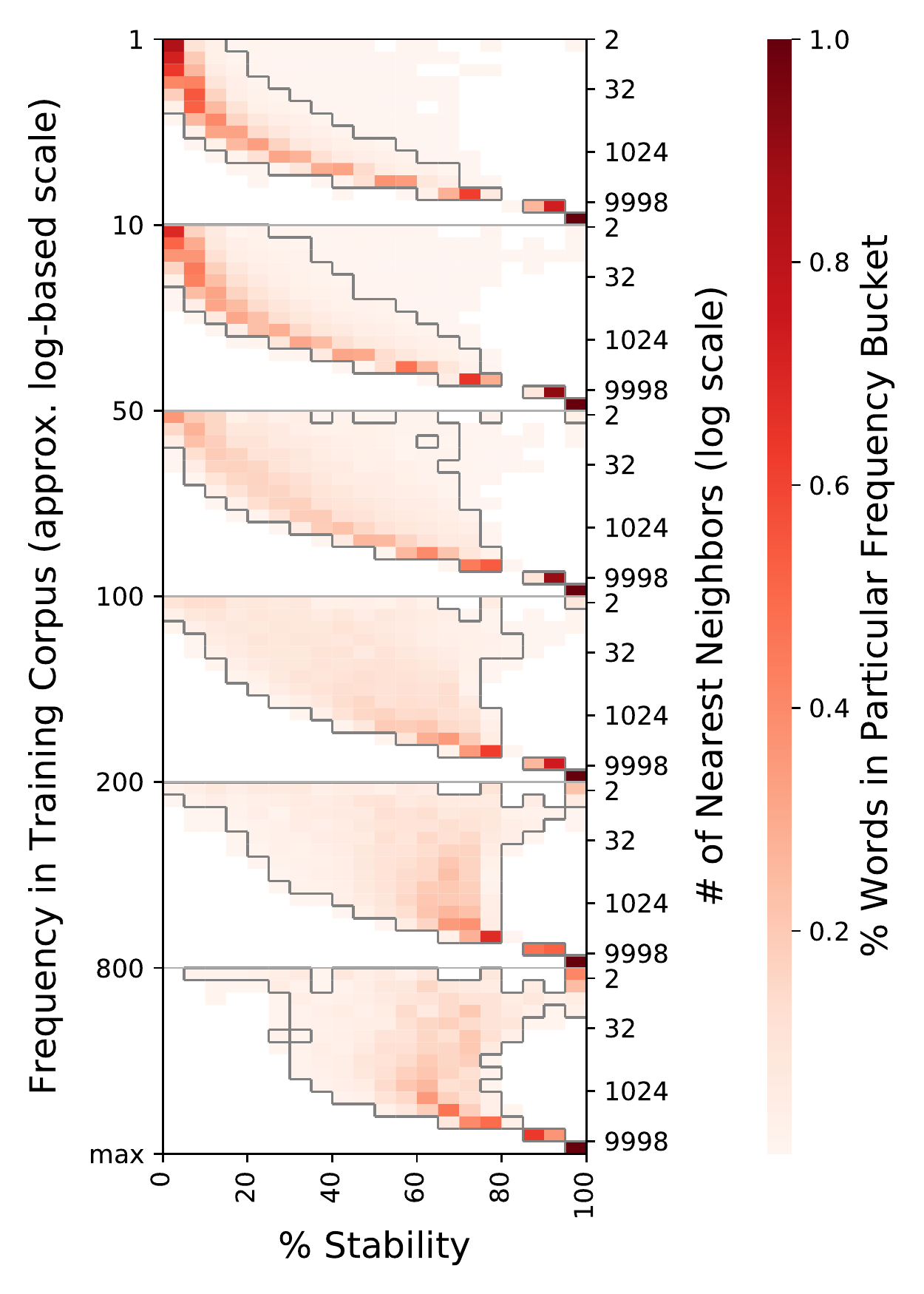}
    \caption{Stability of \textit{word2vec} on the PTB. Stability is measured across ten randomized embedding spaces trained on the training data of the PTB (determined using language modeling splits \cite{mikolov2010recurrent}). Each word is placed in a frequency bucket (left y-axis) and stability is determined using a varying number of nearest neighbors for each frequency bucket (right y-axis). Each row is normalized, and boxes with more than 0.01 of the row's mass are outlined.}
    \label{fig:increasingNeighbors}
\end{figure}

To further compare \textit{word2vec} and GloVe, we look at how the stability of \textit{word2vec} changes with the frequency of the word and the number of neighbors used to calculate stability. This is shown in Figure~\ref{fig:increasingNeighbors} and is directly comparable to Figure~\ref{fig:increasingNeighbors_glove}.
Surprisingly, the stability of \textit{word2vec} varies substantially with the frequency of the word. For lower-frequency words, as the number of nearest neighbors increases, the stability increases approximately exponentially. For high-frequency words, the lowest and highest number of nearest neighbors show the greatest stability. This is different than GloVe, where stability remains reasonably constant across word frequencies, as shown in Figure~\ref{fig:increasingNeighbors_glove}.
The behavior we see here agrees with the conclusion of \cite{mimno2017strange}, who find that GloVe exhibits more well-behaved geometry than \textit{word2vec}.

\mysubsection{Observation 5. Frequency is not a major factor in stability.}
To better understand the role that frequency plays in stability, we run separate ablation experiments comparing regression models with frequency features to regression models without frequency features. Our current model (using raw frequency) achieves an $R^2$ score of 0.301. Comparably, a model using the same features, but with normalized instead of raw frequency, achieves a score of 0.303. Removing frequency from either regression model gives a score of 0.301. This indicates that frequency is not a major factor in stability, though normalized frequency is a larger factor than raw frequency.

Finally, we look at regression models using only frequency features. A model using only raw frequency features has an $R^2$ score of 0.008, while a model with only normalized frequency features has an $R^2$ score of 0.0059. This indicates that while frequency is not a major factor in stability, it is also not negligible. As we pointed out in the introduction, frequency does correlate with stability (Figure~\ref{fig:ptb_stability}). However, in the presence of all of these other features, frequency becomes a minor factor.

\section{Impact of Stability on Downstream Tasks}

Word embeddings are used extensively as the first stage of neural networks throughout NLP.
Typically, embeddings are initalized based on a vector trained with \textit{word2vec} or GloVe and then further modified as part of training for the target task.
We study two downstream tasks to see whether stability impacts performance.

Since we are interested in seeing the impact of word vector stability, we choose tasks that have an intuitive evaluation at the word level: word similarity and POS tagging. %, and language modeling.

\subsection{Word Similarity}

\begin{figure}
    \centering
    \includegraphics[width=0.5\textwidth]{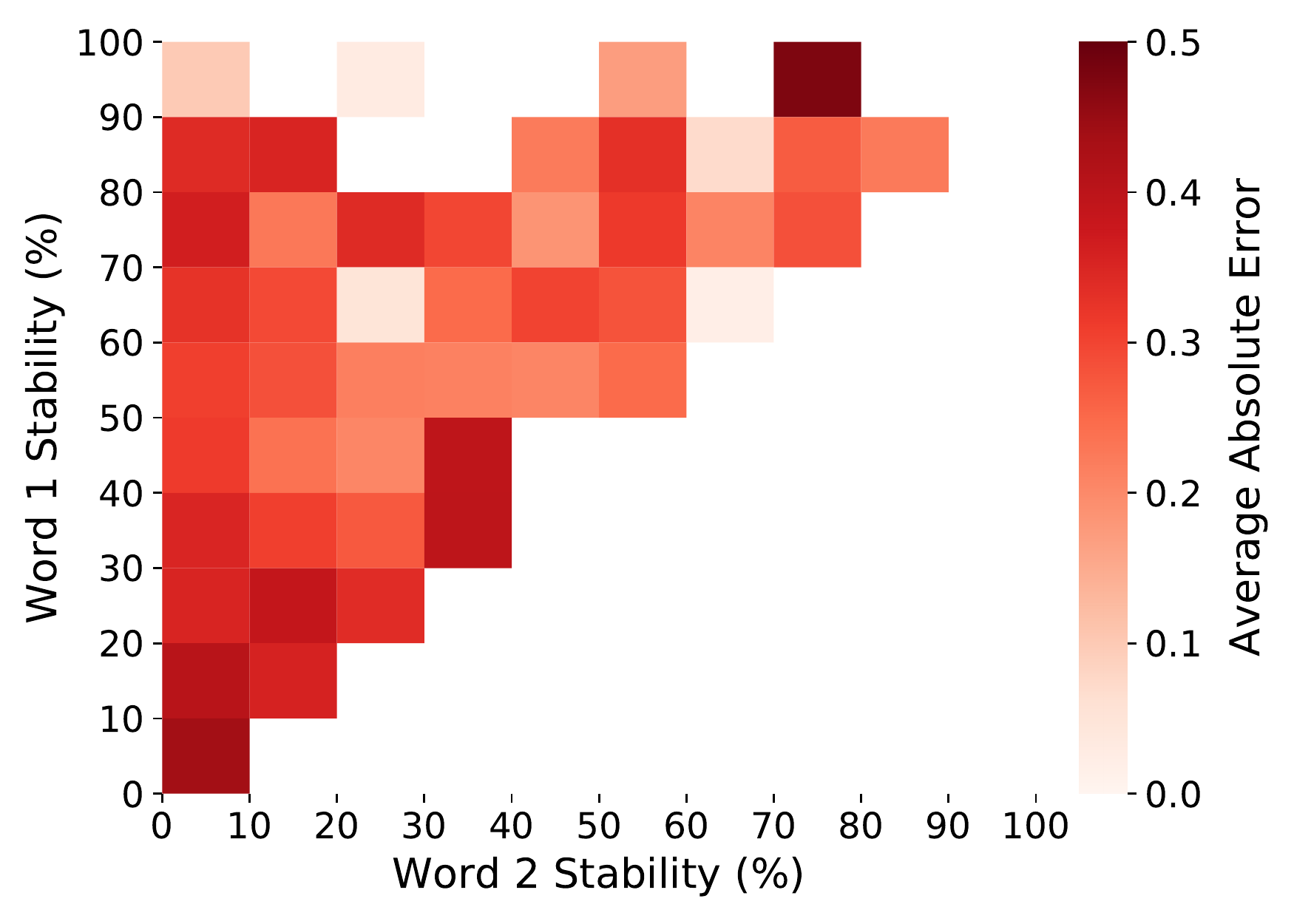}
    \caption{Absolute error for word similarity.}
    \label{fig:individual_wordsim}
\end{figure}

To model word similarity, we use 300-dimensional \textit{word2vec} embedding spaces trained on the PTB. For each pair of words, we take the cosine similarity between those words averaged over ten randomly initialized embedding spaces. %We then normalize this similarity to be in the range $[0,1]$.

We consider three datasets for evaluating word similarity: WS353 (353 pairs) \cite{finkelstein2001placing}, MTurk287 (287 pairs) \cite{radinsky2011word}, and MTurk771 (771 pairs) \cite{halawi2012large}. For each dataset, we normalize the similarity to be in the range $[0,1]$, and we take the absolute difference between our predicted value and the ground-truth value. Figure~\ref{fig:individual_wordsim} shows the results broken down by stability of the two words (we always consider Word 1 to be the more stable word in the pair). Word similarity pairs where one of the words is not present in the PTB are omitted.

We find that these word similarity datasets do not contain a balanced distribution of words with respect to stability; there are substantially more unstable words than there are stable words. However, we still see a slight trend: As the combined stability of the two words increases, the average absolute error decreases, as reflected by the lighter color of the cells in Figure~\ref{fig:individual_wordsim}  while moving away from the (0,0) data point. 

\subsection{Part-of-Speech Tagging}

\begin{figure}[!h]
    \centering
    \begin{subfigure}[b]{0.48\textwidth}
       \includegraphics[trim={0 4mm 0 0},width=1\linewidth]{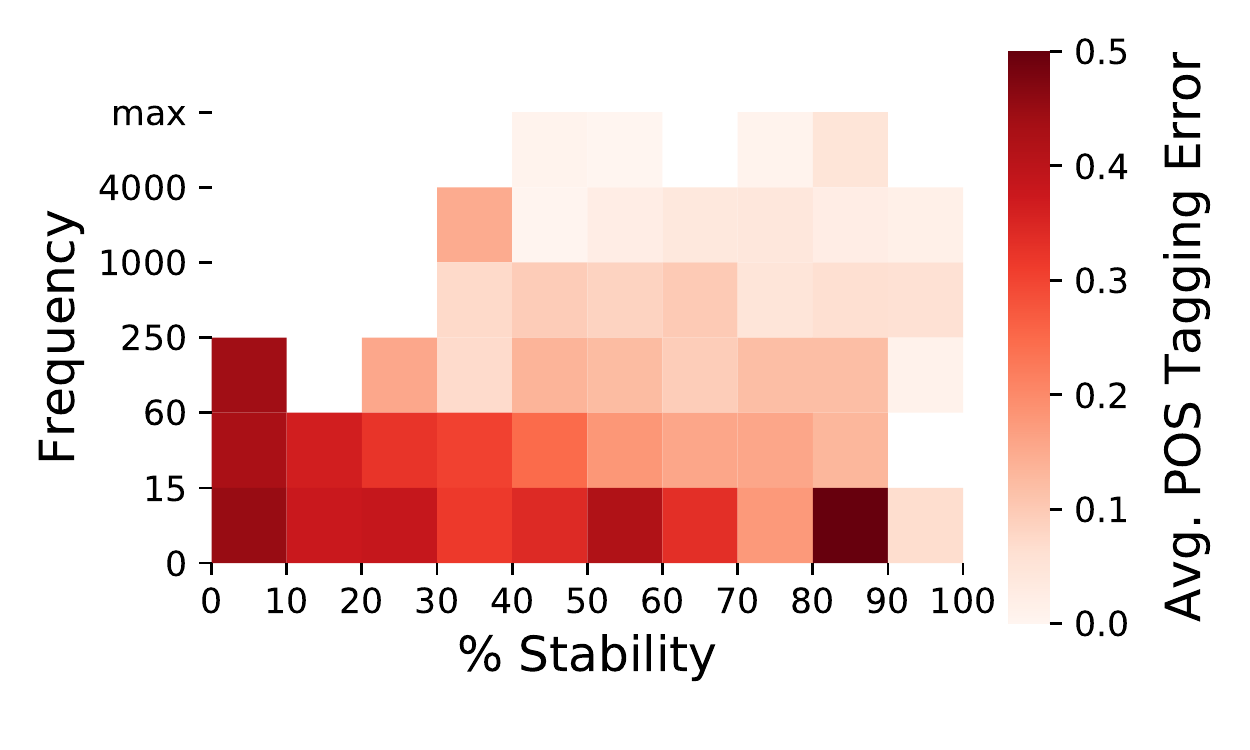}
       \caption{\label{fig:pos-fixed}POS error probability with fixed vectors.}
    \end{subfigure}
    
    %Laura: I changed this to be the same scale as the above graph, which is why they don't look as comparable before. I think this more clearly represents what's going on, but if it's hard to read, I can change it back.
    %JKK: I really like the change!
    \begin{subfigure}[b]{0.48\textwidth}
       \includegraphics[trim={0 4mm 0 0},width=1\linewidth]{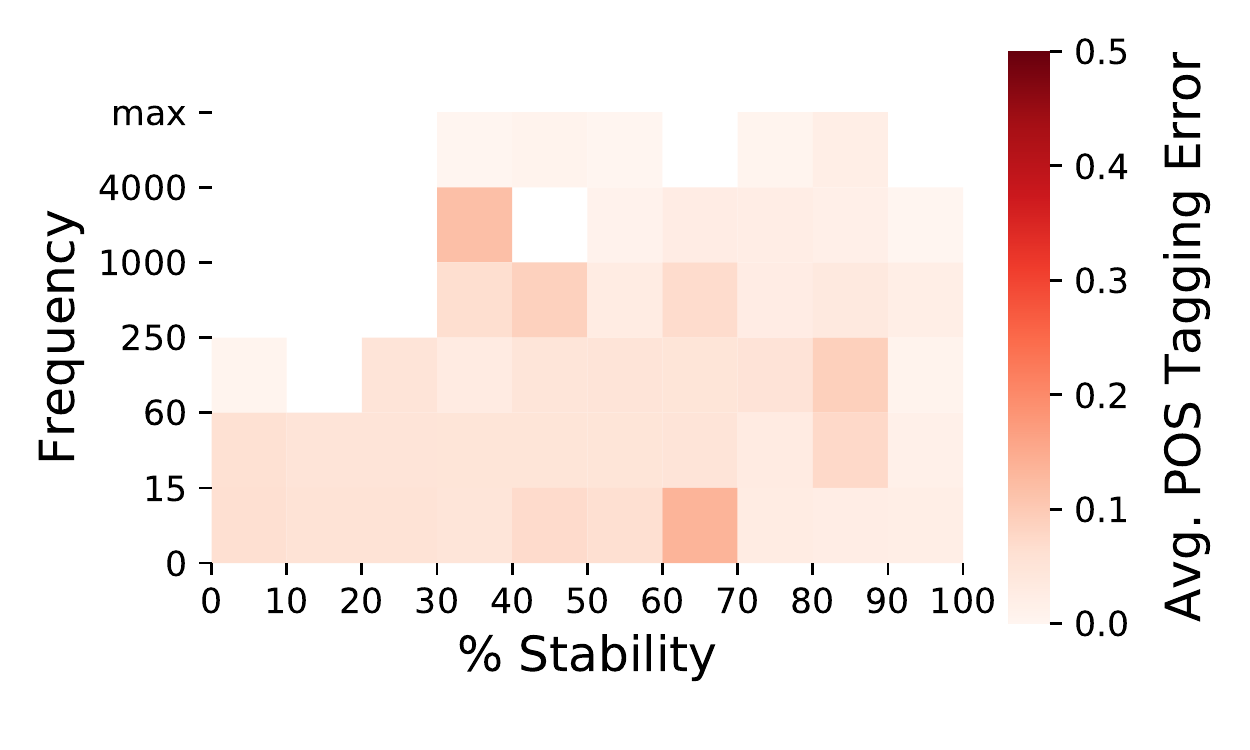}
       \caption{\label{fig:pos-shift}POS error probability when vectors may shift in training.}
    \end{subfigure}
    
    \begin{subfigure}[b]{0.48\textwidth}
       \includegraphics[trim={0 8mm 0 0},width=1\linewidth]{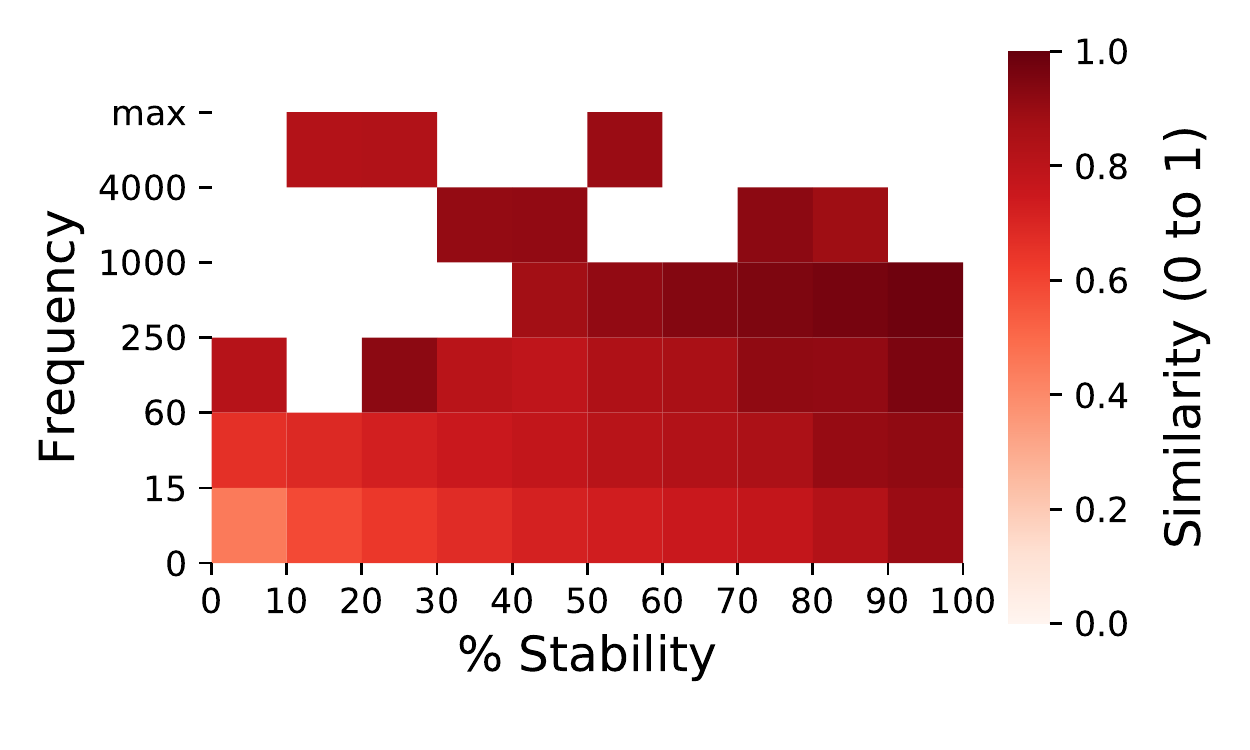}
        \caption{\label{fig:word_vector_shift}
        Cosine similarity between original word vectors and shifted word vectors.}
    \end{subfigure}
    
    \caption{
    \label{fig:pos_stability}
    Results for POS tagging.
    (\subref{fig:pos-fixed}) and (\subref{fig:pos-shift}) show average POS tagging error divided by the number of tokens (darker is more errors) while either keeping word vectors fixed or not during training. (\subref{fig:word_vector_shift}) shows word vector shift, measured as cosine similarity between initial and final vectors. In all graphs, words are bucketed by frequency and stability.}
\end{figure}

Part-of-speech (POS) tagging is a substantially more complicated task than word similarity.
We use a bidirectional LSTM implemented using DyNet \cite{dynet}.
We train nine sets of 128-dimensional word embeddings with \textit{word2vec} using different random seeds.
The LSTM has a single layer and 50-dimensional hidden vectors. Outputs are passed through a \textit{tanh} layer before classification.
To train, we use SGD with a learning rate of 0.1, an input noise rate of 0.1, and recurrent dropout of 0.4.

This simple model is not state-of-the-art, scoring 95.5\% on the development set, but the word vectors are a central part of the model, providing a clear signal of their impact.
For each word, we group tokens based on stability and frequency.
Figure~\ref{fig:pos_stability} shows the results.\footnote{The unusual dark spot that occurs at medium-high stability and low frequency is caused primarily by words that have a substantially different POS distribution in the test data than in the training data.}
Fixing the word vectors provides a clearer pattern in the results, but also leads to much worse performance: 85.0\% on the development set.
Based on these results, it seems that training appears to compensate for stability.
This hypothesis is supported by Figure~\ref{fig:word_vector_shift}, which shows the similarity between the original word vectors and the shifted word vectors produced by the training.
In general, lower stability words are shifted more during training.

Understanding how the LSTM is changing the input embeddings is useful information for tasks with limited data, and it could allow us to improve embeddings and LSTM training for these low-resource tasks.

\section{Conclusion and Recommendations}

Word embeddings are surprisingly variable, even for relatively high frequency words.
Using a regression model, we show that domain and part-of-speech are key factors of instability.
Downstream experiments show that stability impacts tasks using embedding-based features, though allowing embeddings to shift during training can reduce this effect.
In order to use the most stable embedding spaces for future tasks, we recommend either using GloVe or learning a good curriculum for \textit{word2vec} training data. We also recommend using in-domain embeddings whenever possible.

The code used in the experiments described in this paper is publicly available from \url{http://lit.eecs.umich.edu/downloads.html}.

\section*{Acknowledgments}
We would like to thank Ben King and David Jurgens for helpful discussions about this paper, as well as our anonymous reviewers for useful feedback. 
This material is based in part upon work supported by the National Science Foundation (NSF \#1344257) and the Michigan Institute for Data Science (MIDAS). Any opinions, findings, and conclusions or recommendations expressed in this material are those of the authors and do not necessarily reflect the views of the NSF or MIDAS. 

\bibliographystyle{acl_natbib}
\bibliography{references}

\end{document}